\documentclass[journal]{IEEEtran}
\usepackage{graphicx}
\usepackage{amsfonts}

\usepackage[english]{babel}
\usepackage{amsmath, amssymb}
\usepackage{float}
\usepackage{makecell}
\usepackage{multirow}
\usepackage{cite}
\usepackage{amsmath,amssymb,amsfonts}
\usepackage{algorithmic}
\usepackage{graphicx}
\usepackage{textcomp}
\usepackage{xcolor}

\usepackage{subfigure}
\usepackage{epstopdf}

\usepackage{graphicx}
\usepackage{comment}
\usepackage{amsmath,amssymb} 
\usepackage{color}
\usepackage{enumerate}

\usepackage{booktabs}
\usepackage[ruled,vlined]{algorithm2e}

\ifCLASSINFOpdf
\else
\fi
\begin{document}
%
\title{
 Learning of Long-Horizon Sparse-Reward Robotic Manipulator Tasks with Base Controllers
}
%
%
%

\author{Guangming Wang, Minjian Xin, Wenhua Wu, Zhe Liu, and Hesheng Wang
        
\thanks{*This work was supported in part by the Natural Science Foundation of China under Grant 62073222 and U1913204, in part by “Shu Guang” project supported by Shanghai Municipal Education Commission and Shanghai Education Development Foundation under Grant 19SG08,  in part by Shenzhen Science and Technology Program under Grant JSGG20201103094400002, in part by the Science and Technology Commission of Shanghai Municipality under Grant 21511101900, in part by grants from NVIDIA Corporation. The first two authors contributed equally. Corresponding Author: Hesheng Wang.}
\thanks{G. Wang, W. Wu, and H. Wang are with Department of Automation, Key Laboratory of System Control and Information Processing of Ministry of Education, Key Laboratory of Marine Intelligent Equipment and System of Ministry of Education, Shanghai Engineering Research Center of Intelligent Control and Management, Shanghai Jiao Tong University, Shanghai 200240, China.}
\thanks{M. Xin is with the  Department of Computer Science and Engineering at University of California San Diego, La
Jolla, CA 92093, U.S.} 

\thanks{Z. Liu is with the Department of Computer Science and Technology, University of Cambridge, Cambridge, CB2 1TN, U.K.}

}

%
%

\markboth{Journal of \LaTeX\ Class Files,~Vol.~14, No.~8, August~2015}%
{Shell \MakeLowercase{\textit{et al.}}: Bare Demo of IEEEtran.cls for IEEE Journals}
%



\maketitle


\begin{abstract}
Deep Reinforcement Learning (DRL) enables robots to perform some intelligent tasks end-to-end. However, there are still many challenges for long-horizon sparse-reward robotic manipulator tasks. On the one hand, a sparse-reward setting causes exploration inefficient. On the other hand, exploration using physical robots is of high cost and unsafe. In this paper, we propose a method of learning long-horizon sparse-reward tasks utilizing one or more existing traditional controllers named base controllers in this paper. Built upon Deep Deterministic Policy Gradients (DDPG), our algorithm incorporates the existing base controllers into stages of exploration, value learning, and policy update. Furthermore, we present a straightforward way of synthesizing different base controllers to integrate their strengths.
Through experiments ranging from stacking blocks to cups, it is demonstrated that the learned state-based or image-based policies steadily outperform base controllers. Compared to previous works of learning from demonstrations, our method improves sample efficiency by orders of magnitude and improves the performance. Overall, our method bears the potential of leveraging existing industrial robot manipulation systems to build more flexible and intelligent controllers.
\end{abstract}

\begin{IEEEkeywords}
Long-horizon sparse reward, deep reinforcement learning, robotic manipulator tasks, base controllers.
\end{IEEEkeywords}

%
\IEEEpeerreviewmaketitle

\section{Introduction}
\IEEEPARstart{D}{eep} Reinforcement Learning (DRL) has been applied extensively to complicated decision-making problems in the past few years, including board games like Go\cite{silver2017mastering} and video games like StarCraft\cite{vinyals2019grandmaster}. Meanwhile, a lot of research has been done on applying DRL to robot control problems \cite{kiumarsi2017optimal,hu2019reinforcement,li2019deep, andrychowicz2020learning,bai2021addressing,gu2017deep, pushgrasp, kalashnikov2018scalable}. However, DRL algorithms are often data-hungry, requiring millions of interactions with the environment to learn a single task. Exploration is even more inefficient when the reward is sparse and delayed, which is a common situation in robot tasks. Besides the random exploration of a robot in the real world is of high cost and unsafe.
Although there is a scheme of training in simulation and testing in real environment, the gap between simulation and real environment always exists, which requires a high degree of verisimilitude of the simulation environment.

Previous works have used demonstrations \cite{DDPGfD, diverse_visuomotor, adaptivecurriculum, goalconditioned,xiang2019task} to speed up DRL. Although this method could successfully learn some long-horizon sparse-reward tasks, it is still not a perfect solution to the efficiency problem. Learning tasks like ``pick-and-stack" still take millions of interactions with the environment. Furthermore, additional hardware like virtual reality equipment or 3D motion controllers are required to collect demonstrations. 

For the sparse reward task, it is difficult for the robot to complete the task and  obtain the reward by exploring on its own. In that case, the original DDPG does not work or convergence is very slow. Therefore, we propose an algorithm guided by one or more base controllers. The base controller can provide a better start of learning and guide the network controller to learn the advantages of base controllers, making the training of the network-based controller faster and safer. At the same time, the network controller can compensate for the single fixed operation mode of the base controller, so that more advantageous strategies can be explored based on the base controller in the learning process. Although the demonstration-based method can also conduct guidance, its guidance efficiency is relatively low and the agent learns slowly.

Although residual RL \cite{2019residual, schoettler2020deep} also introduces a baseline in the learning process of the network controller, its network only learns the correction of the  base controller. Its final effect is limited by the base controller, so it is difficult to explore a better strategy. It is also difficult for DDPG to find appropriate compensation for sparse rewards, which makes the overall effect of the base controller and network controller still poor. At the same time, the network cannot combine the advantages of multiple base controllers.

Considering these, we propose a hybrid method for robotic task learning with existing controllers, which is referred to as base controllers. The base controller has its own fixed mode, and the network will explore based on behavior model of the base controller. Conservative and slow controllers with high accuracy and aggressive and efficient controllers with low accuracy are difficult to be selected as appropriate guiding baselines. Therefore, we also study an ensemble strategy of base controllers, to merge the advantages of different base controllers. The network can learn a flexible strategy and make a reasonable choice under different base controllers for learning, to get an end-to-end controller with high efficiency and high precision. The main contributions of this paper include:
\begin{itemize}
    \item Long horizons and sparse rewards make it difficult for robots to obtain rewards to optimize networks. We present a novel learning scheme that incorporates base controllers to resolve the challenges of sparse-reward and long-horizon robotic tasks. The proposed approach greatly relieves the exploration cost during learning.
    \item We extend our approach to an ensemble of base controllers so that the network can aggregate the advantages of these controllers. The final network performance is better than all these base controllers.
    \item The experiments demonstrate that, compared with learning from demonstrations, our approach accelerates the learning speed by orders of magnitude and is also effective in handling difficult tasks which can not be solved by demonstration learners.
\end{itemize}

\par The rest of this article is organized as follows. Section II summarizes some works related to this paper. Section III describes our method. Section IV shows the experiment details and results. Finally, the conclusion is drawn in the fifth section.  

\section{Related Work}

\subsection{Imitation Learning}
Imitation Learning (IL) is broadly adopted in robotics, ranging from automated driving \cite{BC, selfdriving} to Unmanned Aerial Vehicle (UAV) control \cite{UAV}. The simplest way of doing IL is Behavior Cloning (BC) \cite{BC}, which intends to match the actions of the learner to those of an expert. BC is not safe due to over-fitting and distribution drift issues. IL algorithms like Dataset Aggregation (DA{\small GGER}) \cite{dagger, menda2019ensembledagger} address this by collecting data with an annealed ratio of expert control, which helps achieve low-cost learning or regret-minimization. Nevertheless, a huge amount of expert data needs to be collected. Recent algorithms can also learn implicitly from demonstration data, such as Inverse Reinforcement Learning (IRL) \cite{ng2000algorithms} and Generative Adversarial Imitation Learning (GAIL) \cite{GAIL}.

\subsection{Reinforcement Learning}
State-of-the-art DRL algorithms have been applied in the robot learning field, ranging from robotic manipulator control using Deep Q-Network (DQN) \cite{DQN, zhang2015towards}, dexterous robot hand manipulation using Proximal Policy Optimization (PPO) \cite{PPO, andrychowicz2020learning}, up to applications of Deep Deterministic Policy Gradient (DDPG) \cite{lillicrap2016continuous, practicalinsertion}, and Soft Actor-Critic (SAC) \cite{haarnoja2018soft}. For simple sparse-reward settings, Hindsight Experience Replay (HER) \cite{HER} proposes the notion of goal-conditioned tasks. But HER is hard to be applied to more complex tasks with bottlenecks \cite{goalconditioned}. DivAC \cite{yang2021maximum} extends the divergent strategy iterative algorithm to the continuous action space problem, deals with the tradeoff between exploration and exploitation, and is able to solve the problem of high-dimensional continuous action space. Yang et al. \cite{yang2021hierarchical} propose the Universal Option Framework (UOF) to solve the multistep tasks by training symbol planning and motion control strategies simultaneously.

\subsection{Combining Imitation With Reinforcement Learning}
Researchers have been trying to combine IL with RL to get the best of both worlds. DDPGGfD\cite{DDPGfD_origin} puts demonstration in the replay buffer of DDPG to learn peg insertion tasks. It uses prioritized replay for sampling transitions across both the demonstration and agent data. Nair et al. \cite{DDPGfD} augment DDPG with a Q-filtered BC loss to learn with a separate demonstration replay buffer and show that the robot is able to learn a multi-step sparse-reward block-stacking task. In our approach, all transformations are stored in the replay buffer of DDPG, and there is no distinction between whether the actions in the transformation are decided by the base controller or the agent. The model is more concise. We gradually reduce the possibility of taking a base controller by a decaying coefficient. Our behavior cloning loss is similar to \cite{DDPGfD}. However, the BC loss of \cite{DDPGfD} is only calculated in the demonstration, and we will calculate the BC loss when the decision of actor network is inferior to the base controller. This allows the actor network to learn strategy from the base controller more quickly and gradually reduce the dependence on the base controller.

Zhu et al. \cite{diverse_visuomotor} employ GAIL to learn from demonstrations, and use state prediction as auxiliary tasks to learn a deep visuomotor policy for a robot hand. Ding et al. \cite{goalconditioned} use demonstrations to overcome bottlenecks in HER. Hermann et al. \cite{adaptivecurriculum} sample demonstrations according to a curriculum of increasing difficulty to help learn a visuomotor policy. All of the above works use demonstrations to realise robot learning, but tasks like ``pick-and-stack" still take millions of environment interactions. Additional hardware are also needed. Specifically, Nair et al. \cite{DDPGfD} use virtual reality equipment, Zhu et al. \cite{diverse_visuomotor} use 3D motion controllers, and Hermann et al. \cite{adaptivecurriculum} use a 3D mouse. Finally, the agent is prone to making mistakes in the initial stage when learning from demonstrations, which is unfavorable when mistake costs are expensive. 
In contrast to learning from demonstrations, learning with existing controllers often directly uses a controller to aid exploration. Guided Policy Search (GPS) \cite{GPS, GPS2} optimizes a model-based controller and then learns a policy in the supervised manner. Residual RL \cite{2019residual, schoettler2020deep} learns a correction policy using RL on top of a feedback controller. One major characteristic of residual RL is that it only learns a compensation policy that is strongly dependent on the existing controller. However, it does not apply to the insertion tasks with sparse rewards discussed in this paper, while our method learns a complete policy and the base controller can be discarded once learning is done. Our method can also utilize an ensemble of base controllers, while residual RL can only use one controller. 

\begin{figure*}[t]
    \centering
    \includegraphics[width=1.0\textwidth]{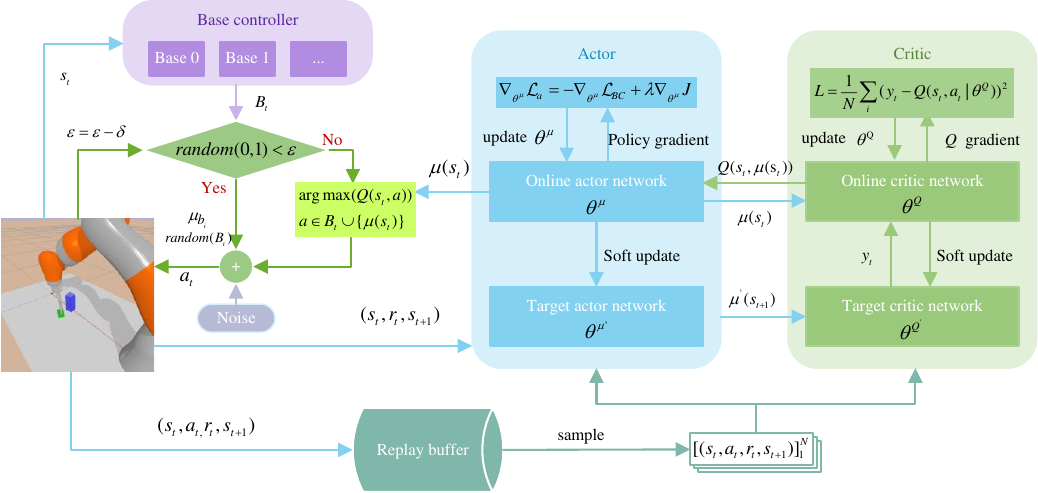}
    \caption{The algorithm overview of DDPG with base controllers (DDPGwB). First, the state information is obtained from the environment. Multiple base controllers are used to get the base controller action set $B_t$. A random action is selected from $B_t$ with probability $\epsilon$. With probability $1-\epsilon$, Equation (\ref{equ:macroensemble}) is used to select the most valuable action from the actions explored by agent and the base controller actions, and then the action is executed with a certain Gaussian noise. Every step $\epsilon$ decays by $\delta$.
    A mini-batch of $N$ transitions is randomly sampled from replay buffer, and $y_t$ is calculated by Equation (\ref{equ:bbensemble}). The critic network is updated, and the actor network is updated by Equation (\ref{equ:la}). Finally, the target networks are soft updated.
    }
    \label{fig:ddpgwB}
\end{figure*}

\section{METHOD}

We formulate the Markov Decision Process (MDP) as standard RL. For state set $\mathcal{S}$ and action set $\mathcal{A}$ in environment $E$, a policy $\mu$ is a mapping $\mathcal{S}\rightarrow\mathcal{A}$, and $\mathcal{P}: \mathcal{S}\times\mathcal{A}\rightarrow\mathcal{S}$ is the transition probability. For every time step $t$, the agent receives observation $s_t$, takes an action $a_t$, and then receives next observation $s_{t+1}$ and reward $r_t$. The goal is to maximize the accumulated discounted reward $R_t=\sum_{t=t}^T\gamma^{i-t}r_t$ where $\gamma$ is the discount factor. The optimal Q-value is defined as the expected return from a state after taking an action and acting optimally thereafter: $Q^*(s_t, a_t)=\max\limits_{\mu}\mathbb{E}_{s_t, r_t\sim E, a_t\sim\mu}[R_t|s_t,a_t]$. A base controller is denoted as $\mu_b(s)$ and can be queried for actions regarding arbitrary states. Our algorithm builds on DDPG \cite{lillicrap2016continuous}, a standard actor-critic algorithm for continuous control which is often preferred in previous works \cite{DDPGfD_origin, DDPGfD} for its utilization of off-policy data. The actor network is denoted as $\mu(s|\theta^\mu)$, and the critic network is denoted as $Q(s, a|\theta^Q)$, where $\theta^\mu$ and $\theta^Q$ are corresponding network weights. In the subsections below, we will describe our algorithm, DDPG with Base Controllers (DDPGwB), as shown in Fig. \ref{fig:ddpgwB}, including methods during exploration, value estimation and policy update to incorporate a base controller. We will also show how the algorithm scales from one base controller to an ensemble of base controllers.\

\subsection{Mixed Q-Control}

Exploration is challenging in robotic tasks, and even more so when the reward is sparse. By utilizing a base controller to explore in the initial stage and linearly decaying its control ratio, the agent is able to experience a regular positive reward. The agent also avoids unsafe states and unnecessary explorations. However, simply annealing the ratio could result in abandoning the base controller too early or too late. When abandoning too early, the agent might crumble due to insufficient guidance, and when abandoning too late, the agent might not outperform the base controller since it cannot explore on its own. To address this, a novel mechanism is proposed that allows the agent to flexibly adjust the control ratio. Specifically, the probability that an agent must execute the base controller's action, $\epsilon$, is 1 at the start of learning and decreases by $\delta$ every environment step. With $1-\epsilon$ probability, the agent selects the best action based on the prediction of critic network:
\begin{equation}
\label{equ:macro}
    a_t = \mathop{\arg\max}_{a\in\{\mu(s_t|\theta^\mu), \mu_b(s_t) \}}Q(s_t, a|\theta^Q).
\end{equation}

The intuition behind this is that in Q-learning \cite{qlearn}, the learned policy is selecting the action with the greatest value over all possible actions given arbitrary state. Since the critic network in DDPG performs the same role as the Q-network in DQN, it is natural to adopt the same fashion of choosing actions via $\mathop{\arg\max}$.

\subsection{Base Controller Bootstrap}

In the update step of critic network, DDPG, like many other RL algorithms, uses the Bellman equation:
\begin{equation}
    Q^*(s_t,a_t) = \mathbb{E}_{s_{t+1}, r_t\sim E}[r_t + \gamma\max\limits_{a_{t+1}} Q^*(s_{t+1},a_{t+1})],
\end{equation}
to estimate the Q-value, where the `$*$' denotes the optimal policy. Unlike Q-learning, DDPG deals with continuous actions and thus loses the ability of efficiently computing the maximum Q-value of next state. Hence, DDPG uses the Q-value of its current policy. In practice, target networks (indicated with `$'$' below) are used for stabilized learning, whose weights are slowly updated. We use $y_t$ to represent the bootstrap target:
\begin{equation}
\label{equ:ddpg}
    y_t = r_t + \gamma Q'(s_{t+1}, \mu'(s_{t+1}|\theta^{\mu'})|\theta^{Q'}).
\end{equation}

 A base controller is adopted to guide exploration in a sparse reward scenario, and the only reward comes at the end of successfully completing the task. Therefore, upon learning the Q-value of that final step, it may not propagate instantaneously backward since the action to reach that reward is very likely different from the agent's own policy at the start. This would possibly contribute to slow-downs and inaccurate Q-value estimates. Hence we propose taking a maximum operation over the Q-values of the agent's own policy and base controller's policy to compose the bootstrap target:
\begin{align}
\label{equ:bb}
    y_t = r_t + \gamma \mathop{\max}[&Q'(s_{t+1}, \mu'(s_{t+1}|\theta^{\mu'})|\theta^{Q'}),\nonumber\\ &Q'(s_{t+1}, \mu_b(s_{t+1})|\theta^{Q'})].  
\end{align}

\begin{algorithm}[t]
\LinesNumbered
\label{DDPGwB}
\caption{DDPG with Base Controllers (DDPGwB)}
 \KwIn{actor network $\mu(s|\theta^\mu)$, critic network $Q(s,a|\theta^Q)$, target network $\mu'$ and $Q'$, $K$ base controllers: $\mu_{b_1}(s), \mu_{b_2}(s), ..., \mu_{b_K}(s)$, replay buffer $R$, Gaussian noise $\mathcal{N}$}
 \KwOut{learned policy $\mu(s|\theta^\mu)$}
Initialize $\epsilon=1$\;
\For{episode={\rm 1,~M}}
{
    Get initial observation $s_1$\;
    \For{t={\rm 1,~T}}
    {
        \If{$random < \epsilon$}
        {
          $a_t= \mu_{b_i}(s_t)$ $random B_{t}$\;
        }
        \Else
        {
          Get $a_t$ according to Equation (\ref{equ:macroensemble});
        }
        $a_t \leftarrow a_t+\mathcal{N}$\;
        Execute action $a_t$, observe reward $r_t$ and next state $s_{t+1}$\;
        Store transition $(s_t,a_t,r_t,s_{t+1})$ in $R$\;
        $\epsilon \leftarrow \epsilon-\delta$\;
        Sample a random minibatch of $N$ transitions $(s_t,a_t,r_t,s_{t+1})$ from $R$\;
        Get $y_t$ according to Equation (\ref{equ:bbensemble})\;
        Update critic by minimizing the Bellman error\;
        Update actor according to Equation (\ref{equ:la})\;
        Update the target networks:
        $\theta^{\mu'} \leftarrow \tau\theta^\mu+(1-\tau)\theta^{\mu'}$,
        $\theta^{Q'} \leftarrow \tau\theta^Q+(1-\tau)\theta^{Q'}$\;
    }
}
\end{algorithm}

This is referred to as base controller bootstrap, combined with mixed Q-control, can also be viewed as learning with a macro policy in the form of Equation (\ref{equ:macro}). Equation (\ref{equ:bb}) can be obtained by plugging the macro policy into the standard bootstrap target form as Equation (\ref{equ:ddpg}).

\subsection{Regularized Policy Update}

The policy update in DDPG is done by back-propagating through the critic network:
\begin{equation}
\label{equ:dpg}
    \nabla_{\theta^\mu}J = \frac{1}{N}\sum_{i}\nabla_{a}Q(s,a|\theta^{Q})|_{a=\mu(s|\theta^{\mu})}\nabla_{\theta^\mu}\mu(s|\theta^{\mu}).
\end{equation}

Due to inaccurate value estimates and a deterministic policy, such unbounded update tends to be very brittle as discussed in \cite{fujimoto2018addressing}, as the actor network will often try to exploit peaks in the Q-function and diverge. To relieve this, we apply behavior cloning with respect to the base controller as a regularization tool that guides the gradient update direction:
\begin{equation}
\label{equ:la}
    \nabla_{\theta^\mu}\mathcal{L}_a = - \nabla_{\theta^\mu}\mathcal{L}_{BC} + \lambda\nabla_{\theta^\mu}J,
\end{equation}
where $\lambda$ is a coefficient balancing two losses. Specifically, the behavior cloning gradient is imposed only on actions that are worse than those of the base controller as decided by the critic network. Note that this does not violate the optimization goal of maximizing the accumulated reward. This kind of filtering is similar to \cite{DDPGfD}, except in \cite{DDPGfD} BC loss is imposed on a demonstration buffer. 

\subsection{Ensemble of Base Controllers}
For complex tasks, there is sometimes no good base controller, but some weak controllers can be easily designed. These controllers have their own characteristics, some high efficiency but low precision, some high precision but low efficiency.
Then, it is ideal to learn a policy from an ensemble of multiple base controllers, and integrate their strengths. To this end, our algorithm easily scales to multiple base controllers. Considering $K$ base controllers: $\mu_{b_1}(s), \mu_{b_2}(s), ..., \mu_{b_K}(s)$, we define the ensemble base controller action set:
\begin{equation}
\label{equ:ensembleactionset}
    \mathcal{B}_t = \{\mu_{b_1}(s_t), \mu_{b_2}(s_t), ..., \mu_{b_K}(s_t)\}.
\end{equation}

With probability $\epsilon$, the action is randomly selected from $B_{t}$. With probability $1-\epsilon$, the choice is made amount actions in $B_t$ and the actor network action according to the modified Equation (\ref{equ:macro}):
\begin{equation}
\label{equ:macroensemble}
    a_t = \mathop{\arg\max}_{a\in\mathcal{B}_t\cup\{\mu(s_t|\theta^\mu)\}}Q(s_t, a|\theta^Q).
\end{equation}
The bootstrap target form in value learning now also extends from Equation (\ref{equ:bb}) to:
\begin{align}
\label{equ:bbensemble}
    y_t = r_t + \gamma \mathop{\max}[&Q'(s_{t+1}, \mu'(s_{t+1}|\theta^{\mu'})|\theta^{Q'}),\nonumber\\ & \mathop{\max}_{a\in\mathcal{B}_{t+1}}Q'(s_{t+1},a|\theta^{Q'})].
\end{align}
Finally, $\mathcal{L}_{BC}$ in Equation (\ref{equ:la}) becomes:
\begin{equation}
\label{equ:lbc}
    \mathcal{L}_{BC} = \frac{1}{N}\sum_{i}||\mathop{\arg\max}_{a\in\mathcal{B}_t\cup\{\mu(s_t|\theta^\mu)\}}Q(s_t, a|\theta^Q) - \mu(s_t|\theta^{\mu})||^2.
\end{equation}

Using an ensemble of base controllers helps the agent to avoid mistakes caused by a single controller during exploration, as in Equation (\ref{equ:macroensemble}). It also helps the critic network to further benefit from a diverse strategies, as reflected in the bootstrap target form in Equation (\ref{equ:bbensemble}). In turn, those learned values help the agent to shape a better policy as in Equation (\ref{equ:lbc}). In this way, even if individual base controller is weak, a learned policy absorbs the strengths in each of them and have a good performance.

The complete pseudo-code is presented in Algorithm \ref{DDPGwB}. We unified exploration, value learning, and policy learning into a single framework to take the best advantage of one or more base controllers.
Base controller bootstrap helps the critic to avoid suffering from a terrible actor, so, the critic network in charge of both exploration and policy update can be learned better. The brittleness of actor network is greatly eased, as it will be shown in experiments that abandoning the use of the target actor network does not create drastic decay to our algorithm’s performance, which is unusual for DDPG.

\section{EXPERIMENTS}

In this section, we first provide a description of the tasks we evaluate our algorithm on. Then the neural network architectures are described, along with a straightforward template for designing base controllers. Finally, we give a quantitative evaluation of our algorithm in terms of learned policies' success rates, learning efficiency, and the ability to utilize an ensemble of base controllers. And we compare to previous methods.

\subsection{Task Description}\label{tasks}

We use PyBullet \cite{pybullet} simulator and a KUKA robotic manipulator for our experiments. In all tasks, the action dimension is 5, consisting of a 3-dimensional (3-D) change of the end-effector position, 1-D change of end-effector yaw angle, and 1-D change of the gripper open angle. All actions have the range of $(-1, 1)$. The observation consists of proprioceptive data of the robotic manipulator (8-D, including end-effector position, orientation, gripper open angle, and the force on gripper fingers), and either state information of two objects (positions and orientations, 12-D) or images (binocular RGB images from two sides of the table, $128\times128$ resolution). Our algorithm is verified under the following three tasks:

\paragraph{Stacking}This task involves two cubes initialized at random positions and orientations on the table, each with a side length of $5cm$. The goal is to stack the green cube onto the purple cube.

\paragraph{Block-cup}This task involves a cube and a cup initialized at random positions and orientations on the table. The cube has a side length of $5cm$. The cup's height is $15cm$ and has a square opening with a side length of $6cm$. The goal is to place the green cube into the blue cup while not tipping the cup over.

\paragraph{Cup-cup}This task involves two cups initialized at random positions and orientations on the table. The green cup's opening has a side length of $5cm$. The blue cup's opening has a side length of $6cm$. Both cups' height is $15cm$. The goal is to place the green cup into the blue cup while not tipping either cup over.

\begin{figure}[t]
  \centering
  \subfigure[Stacking]{\includegraphics[width=0.15\textwidth]{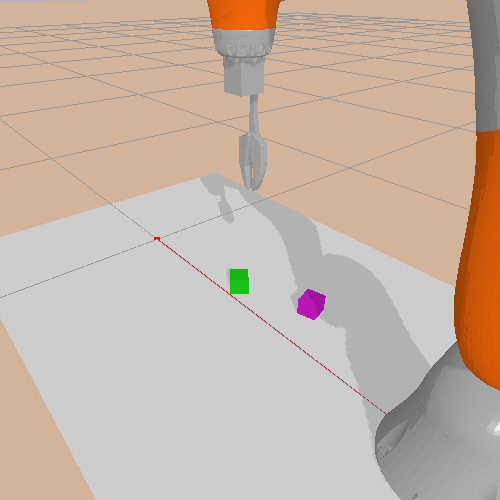}}
  \subfigure[Block-cup]{\includegraphics[width=0.15\textwidth]{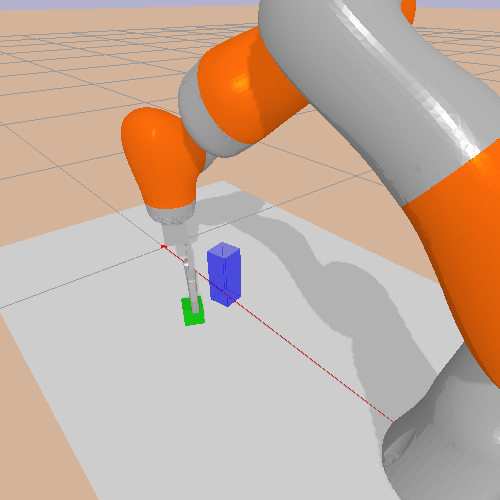}}
  \subfigure[Cup-cup]{\includegraphics[width=0.15\textwidth]{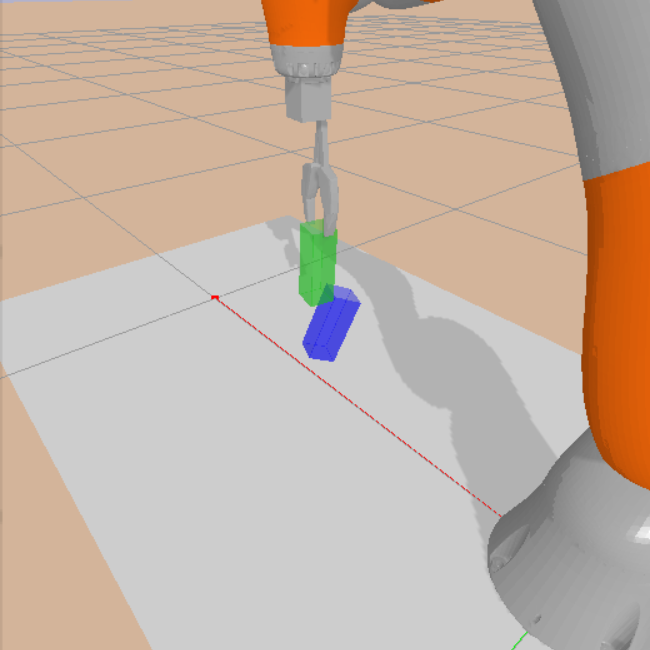}}
  \caption{Screenshots of all tasks in different stages described in Sec. \ref{tasks}. The screenshot of the cup-cup task depicts a failed trial, where the blue cup is tipped over by the robot.}
  \label{fig:task}
\end{figure}

Screenshots of all tasks are presented in Fig. \ref{fig:task}. In all tasks, we employ a truly sparse reward, that is, a reward of 1 if the goal is achieved, and a reward of 0 for other scenarios. An episode is terminated if it reaches 100 steps or the agent achieves the goal. In block-cup or cup-cup tasks, an episode is also terminated when a cup is tipped, and the reward is 0. All tasks are designed to reflect the challenges of sparse-reward and long-horizon robotic tasks, that is, the robot must accomplish a sequence of actions before receiving a proper reward signal. For example, for the cup-cup task, the robot must first reach for the green cup, grasp the cup, then align the green cup with the blue cup, and finally place it into the blue cup to obtain the reward. 

We highlight the increasing difficulties in these tasks: First, the agent has to pick up the object. Second, it cannot tip the cup over when moving the object. Finally, it has to match the object precisely with the opening. The risk of tipping over the cup makes the tasks significantly harder and correspond to a kind of safety requirements, e.g., fragile things are not allowed to be knocked over when the robot is operating. A proper learning scheme should avoid making those expensive mistakes.

\begin{figure*}[t]
    \centering
    \includegraphics[width=1.0\textwidth]{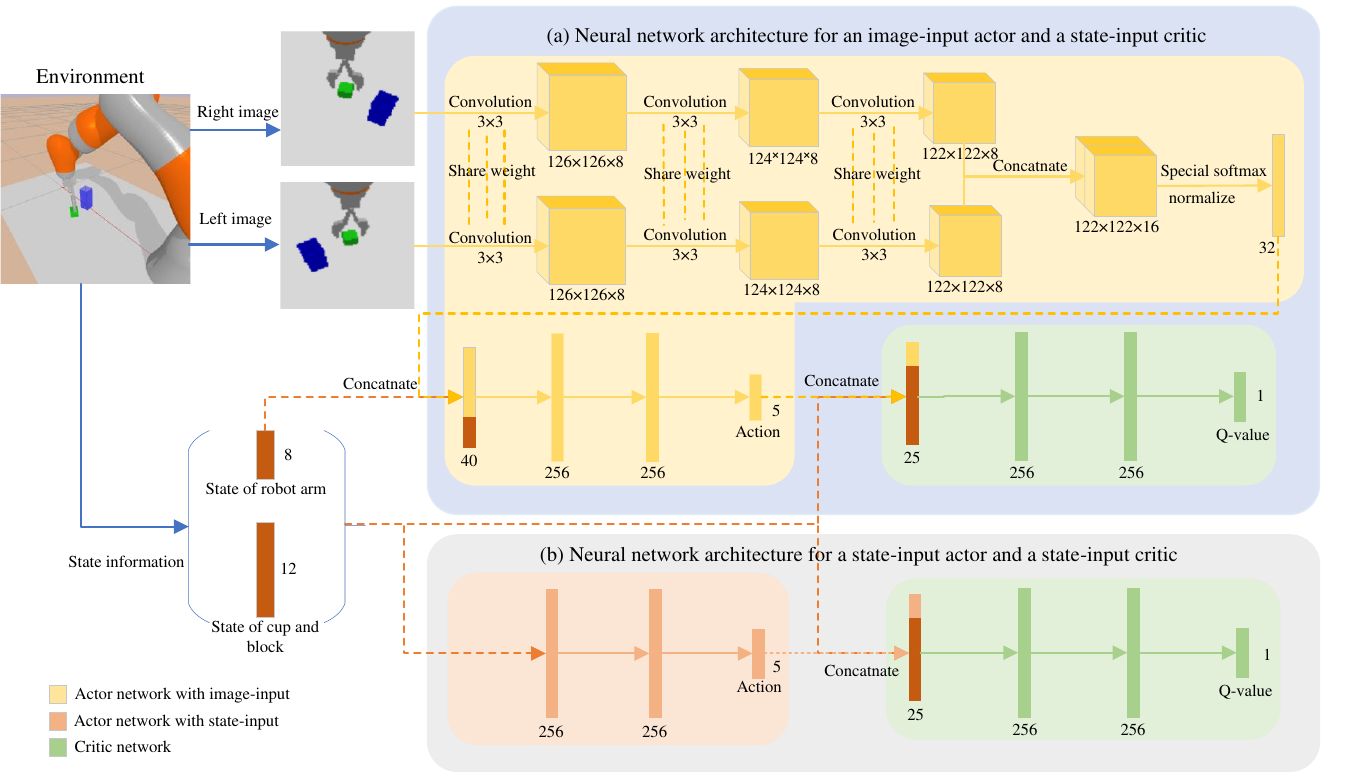}
    \caption{(a) Neural network architecture for an image-input actor and a state-input critic.(b) Neural network architecture for a state-input actor and a state-input critic. (a) is used for image-based policy learning, and (b) is used for state-based policy learning. The state-input critic network in (a) and (b) have the same structure.
    }
    \label{fig:nn}
\end{figure*}

\begin{algorithm}[t]
\LinesNumbered
 \label{template}
 \caption{Base Controller Template}
 \KwIn{\texttt{function} MoveAndAlign: Execute $\textbf{a} = \tanh{\big(}K_p(\textbf{p}_{desired} - \textbf{p}_{curr}){\big)}$}
 \While{goal not achieved}{
 MoveAndAlign\;
 \If{close enough}{
    Close/Open Grip\;
    \If{in grasp}{
    Switch $\textbf{p}_{desired}$\;}
    }
 }
\end{algorithm}

\subsection{Network Structure}

\paragraph{State-input}The actor network receives 20-D input, and outputs 5-D action, with 2 hidden layers each having 256 units and ReLU activation. Activation of the last layer is Tanh. The critic network receives 25-D input and outputs 1-D Q-value, with 2 hidden layers each having 256 units and ReLU activation. Activation of the last layer is Sigmoid.

\paragraph{Image-input}The normalized image first passes through a Convolutional Neural Network (CNN) with 3 convolution layers of $3\times3$ kernel size, 1 stride, 8-channel output, and ReLU activation, then passes through channel-wise spacial softmax \cite{GPS} to get feature points (16-D). The binocular images share one CNN, and their output features are concatenated, resulting in a 32-D feature output. This image feature concatenated with proprioceptive data is then passed through a fully connected network same as the state-input actor network, except for different input dimensions. A state-input critic is coupled with an image-input actor during training as in \cite{diverse_visuomotor}. Note that the use of state information here is for learning only, and the trained image-based policies works without state information. Fig. \ref{fig:nn} outlines the entire neural network architecture. 

\begin{table*}[t]
  \caption{Success rates of test performance and training performance. Test performance is the success rate of actor network after training. 
Training performance is the average success rate of the agent in the interaction process with environment throughout the training process}
  \label{table1}
  \centering
  \setlength{\tabcolsep}{3.mm}{

  \begin{tabular}{lcccccc}
    \toprule
    & & Test performance & & & Training performance &\\
     \cmidrule(lr){2-4}
     \cmidrule(lr){5-7}
    Method & Stacking & Block-cup & Cup-cup & Stacking & Block-cup & Cup-cup \\
    \midrule
  
    HER   \cite{HER}  & 0\% &0\% & 0 \% &0\% & 0\% & 0\%                                \\
 
    DDPG w/BC + Demo \cite{DDPGfD}     & 63.40\%    & 56.90\%  &14.16\%  & 15.79\%& 7.69\% & 1.67\%\\
       Residual RL \cite{2019residual}   & 93.82\% &70.62\% &40.94\% & 82.82\%&57.07\% &21.03\%\\
    Base Controller     & 89.04\%   & 57.66\%    & 28.14\%  & \textbf{89.04\%} &57.66\% & 28.14\%\\
    DDPGwB (Ours)   & \textbf{95.58\%} & \textbf{84.94\%} & \textbf{65.40\%} &87.13\% & \textbf{74.34\%}& \textbf{45.02\%} \\
    \bottomrule
  \end{tabular}}
\end{table*}

\begin{figure*}[t]
  \centering
  \subfigure[]{\includegraphics[width=0.32\textwidth]{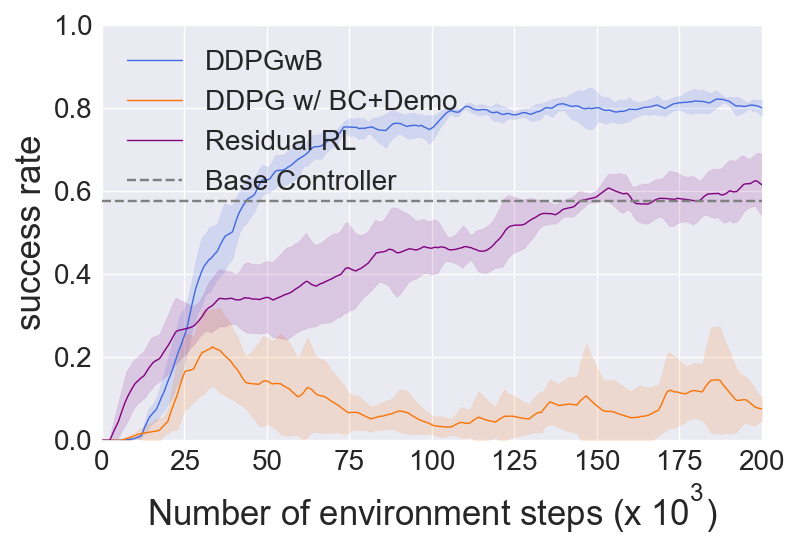}}
   \subfigure[]{\includegraphics[width=0.32\textwidth]{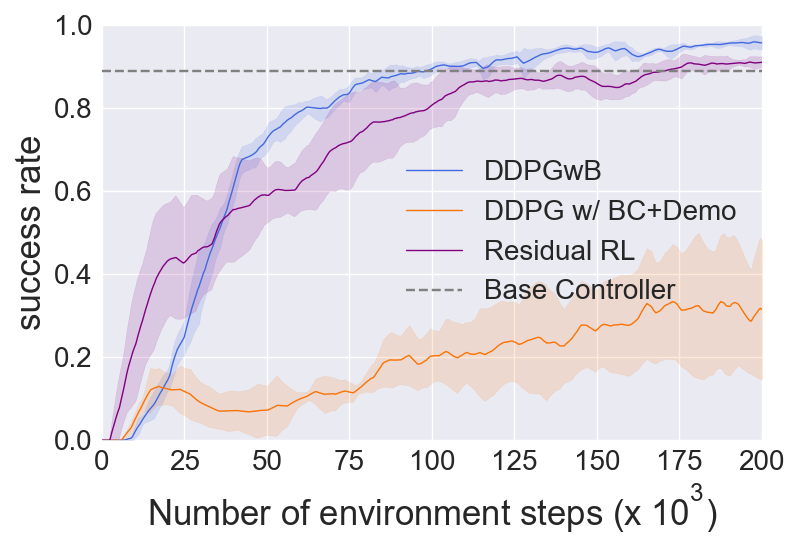}}
    \subfigure[]{\includegraphics[width=0.32\textwidth]{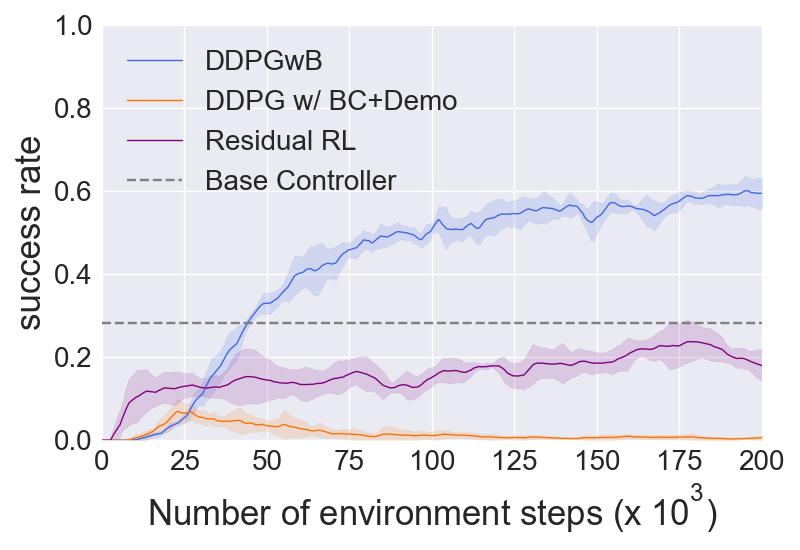}}
  
  \subfigure[]{\includegraphics[width=0.32\textwidth]{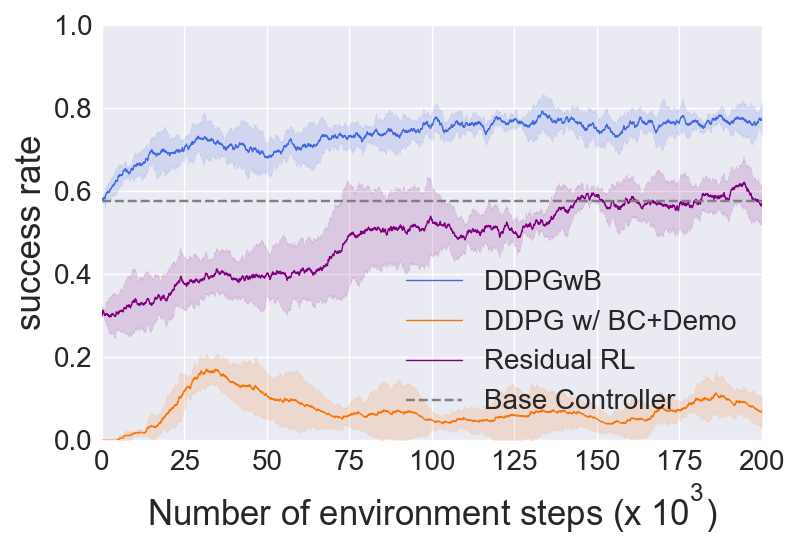}}
  \subfigure[]{\includegraphics[width=0.32\textwidth]{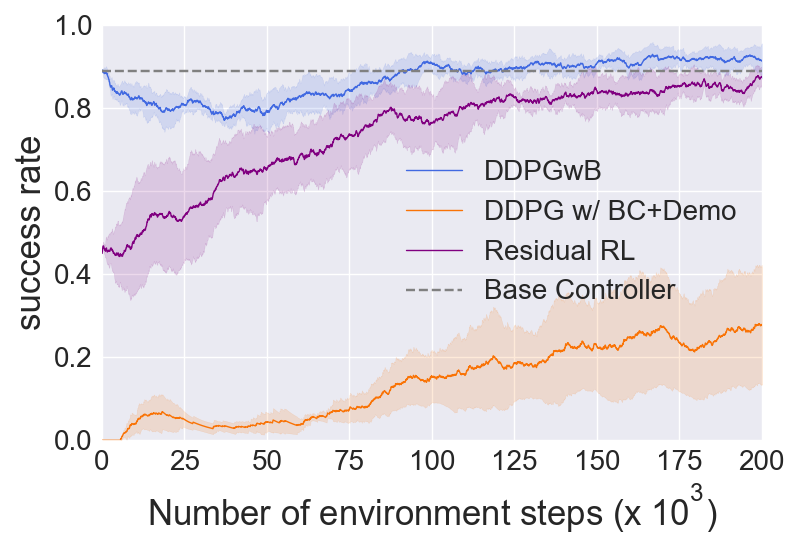}}
  \subfigure[]{\includegraphics[width=0.32\textwidth]{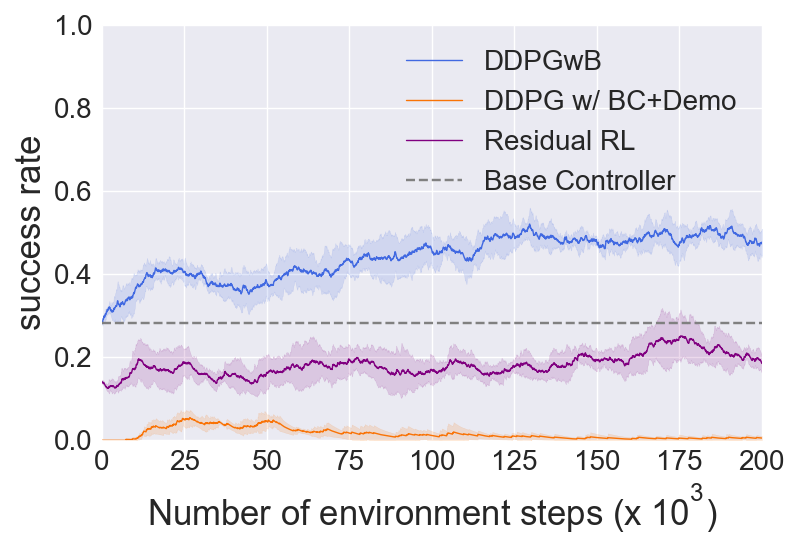}}
  \caption{(a, b, c) Learning curves (success rates during evaluation) of three tasks. (d, e, f) Success rates in actual training of three tasks. Solid lines represent the mean, and shaded areas represent standard deviation. The number of environment steps is used as an x-axis label to convey sample complexity. Curves are averaged from 5 runs with different random seeds and smoothed for clarity. The performance of  base controller is denoted by the horizontal dashed line.
}
  \label{fig:lc}
\end{figure*}

\subsection{Base Controllers}
 Given state information and proprioceptive data, we provide a simple controller template in Algorithm \ref{template} that covers a variety of tasks. $\textbf{p}_{desired}$ represents the desired end-effector (object) position and yaw angle. $\textbf{p}_{curr}$ represents the current end-effector (object) position and yaw angle. The controller first tries to reach and grasp an object. Once the gripper holds the object the controller will place it to the goal position. We use simple proportional controllers to calculate actions for the end-effector. The proportionality coefficient $K_p$ is 5 for position and 2 for angular actions respectively. Force on gripper fingers is used to tell whether there is a successful grasp. The experiments show that Algorithm  \ref{template} produces decent success rates.

\subsection{Training Details}
We use Adam optimizer and a learning rate of $10^{-3}$ for both actor and critic networks. Exploration noise during training is Gaussian distribution with 0.1 standard deviation and 0 mean. Discount factor $\gamma$ is 0.99. Soft update parameter $\tau$ for target networks is $5\times10^{-3}$. Batch size is 256, and replay buffer size is $10^5$. We set $\lambda$ as $2\times10^{-2}$ and $\delta$ as $2\times10^{-5}$. All experiments are performed on a RTX2080 Ti.

\subsection{Results}

During the training, the actor model is tested for 100 episodes every 30 training episodes. We calculate the success rate, and save the model with the highest success rate. After the completion of the training, the saved model is tested for 1000 episodes (no Gaussian noise). The experiments are repeated five times for each experimental condition, and the average of the test results is as the final test performance, listed in Table \ref{table1}. We also measure the training performance, the average success rate of the agent in the interaction process with environment throughout the training process. This reflects the safety requirement of the training process. The results are the training performance in Table \ref{table1}. Due to the Gaussian noise in the training process and the poor performance of the agent in the early training, the performance of the training is worse than that of the test.  

\paragraph{Comparison with Other Methods}
We compare our method with HER\cite{HER}, Residual RL\cite{2019residual}, and the method of learning from demonstrations (DDPG w/BC + Demo) \cite{DDPGfD}. Learning curves (success rates for evaluation) with state-input are summarized in Fig. \ref{fig:lc}(a, b, c). 

For sparse reward tasks, HER\cite{HER} uses a strategy to sample an set of additional goals for replay buffer to improve sampling efficiency. In our tasks, whether the robotic manipulator can successfully grasp the block is one of the bottlenecks of the task. If HER can successfully grasp the block through exploration, HER can strengthen the grasping behavior and start the exploration in the next stage. However, random exploration is almost impossible due to the high complexity of the action space. In their paper, they set half training episodes to start with the block being  grasped. This initialization is equivalent to artificially crossing a bottleneck and making the problem easier. However, this approach is not useful when it is difficult to find one such key frame. Instead, our approach uses the base controller to guide agents to explore, and agents can quickly learn how to grasp. Since we use Mixed Q-Control to choose the optimal execution between the base controller and network actions, most of the transitions in the replay buffer work, which makes our sampling efficiency much higher.

For the method of learning from demonstrations\cite{ DDPGfD}, a total of 500 successful demonstrations are collected via the base controller to compose a demonstration buffer (5 times larger than the buffers in \cite{ DDPGfD}). The demonstration buffer is then sampled during every gradient update, with a Q-filtered BC loss imposed on the policy, as per \cite{DDPGfD}. In all tasks, our method requires only about 0.2 million environment steps or less to converge, while learning from demonstrations struggles to converge in limited environment steps. In previous works of learning sparse-reward tasks from demonstrations, \cite{DDPGfD} learn a much easier pick-and-place task: the robotic manipulator has to pick up a block on the table and move it to a position in space, where neither orientation nor object-interaction is considered. In their paper, \cite{DDPGfD} uses about 1 million environment steps to converge. According to our results, as the difficulty level of the task increases (more bottlenecks), it becomes harder for demonstration learners to find a sensible policy, and performances even start to degrade as the training progresses. For the cup-cup task, the final performance of learning from demonstrations is almost down to zero. 

Residual reinforcement learning\cite{2019residual} also uses base controller. They learn the compensation amount of a base controller and superimpose the compensation amount on the base control action as the actual control action. Although this method can improve the performance of the base controller, the policy is strongly dependent on the base controller. Therefore, other more effective policies cannot be explored, which limits the performance improvement. In addition, for the long-distance sparse reward problem, it is difficult to explore effective compensation for random exploration, so the strategy convergence is slow. In our approach, thanks to Mixed Q-Control and Base Controller Bootstrap, poor base controller performance is slowly discarded as network performance is progressively improved under guidance. Our method will learn the advantages of the base controller and explore a better strategy to replace the poor part of the base controller. This is more flexible and effective than Residual RL.

To characterize the safety of learning, we present the success rates during training in Fig. \ref{fig:lc}(d, e, f), which are the performances the agents hold while actually interacting with the environment. Note that all policies' performances are naturally lower than those during evaluation because of Gaussian noise. Results show that our method learns at a stable success rate throughout the training process. The base controller helps prevent the agent from making mistakes like tipping over the cup in the initial stage. However, Learning from demonstrations start learning with a success rate of 0. At the beginning of training, due to the random compensation of the network, the performance of Residual RL is lower than that of the base controller, and it makes more mistakes. By contrast, our method is safer during training and more suitable for real-world training.

\begin{table}[t]
  \caption{Success rates of test performance}
  \label{table2}
  \centering
  \setlength{\tabcolsep}{0.8mm}{
  \begin{tabular}{lcccc}
    \toprule
    Method &  & Average steps & Success rate & Misbehavior rate  \\
    \midrule  
&  Base 0    & 93.4  & 63.8\%   & 25.9\% \\
    \multirow{-2}{*}{\begin{tabular}[c]{@{}c@{}}Base controller \end{tabular}} & Base 1     & 120.2    & 83.1\%   &13.2\%\\  
   \midrule
   &  Base 0  & \textbf{70.7} & 75.6\% &18.0\% \\
    DDPGwB &Base 1 &96.7 & 82.0\% &14.0\%\\
    &Base 0\&1  &88.3  & \textbf{86.5\%}& \textbf{9.6\%} \\
    \bottomrule
  \end{tabular}}
\end{table}

\paragraph{Learn from Images} We run a set of experiments that use binocular RGB images as actor network's input instead of state information. We train the agent for 0.4 million environment steps (50 gradient steps per episode), and evaluate the final learned policies’ performances by running 1000 test episodes for each task. The image-based policies achieve success rates of 87.28\%, 68.74\%, and
49.18\% respectively for stacking, block-cup and cup-cup task,
proving that our method also applies to image-based tasks. The performance is worse than the state-based result, which we believe is because the feature extraction capability of the convolutional neural network processing images is limited.

\paragraph{Learn with an Ensemble of Base Controllers} 

For an ensemble of base controllers, we design two base controllers, base controller 0 with high efficiency but low accuracy, and base controller 1 with high accuracy but low efficiency. When the ensemble of base controllers works, we randomly choose one of the actions to perform. Firstly, we train our algorithm with these two base controllers respectively. Then, we put the two base controllers into the controller set and train our algorithm with the ensemble of base controllers. The training step length is 1 million steps, and to fully perform the slower base controller, the upper step length of each episode is up to 140 steps.

\begin{figure}[t]
  \centering
  \subfigure{\includegraphics[width=0.42\textwidth]{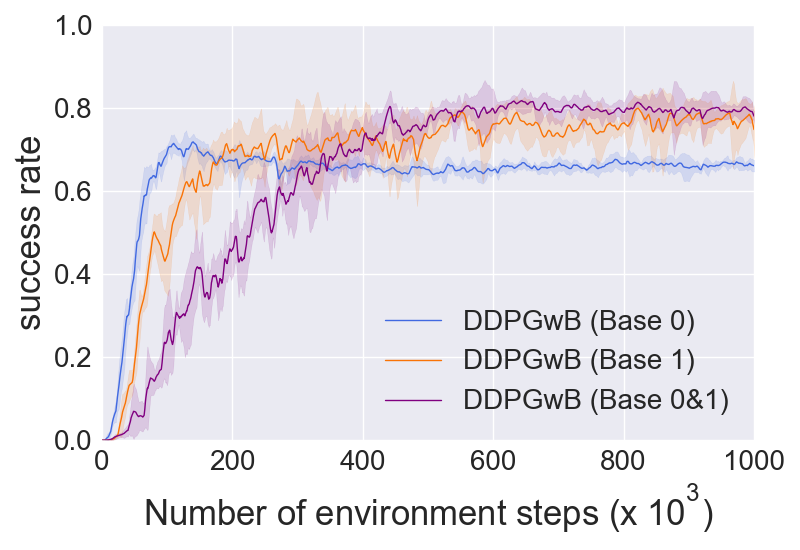}}
  \caption{ Learning curves when using an ensemble of base controllers. 
  Using an ensemble of base controllers ends up converging better than using a single base controller.
  }
  \label{fig:ensemble}
\end{figure}

The experimental results are shown in Fig. \ref{fig:ensemble} and Table \ref{table2}. When the two base controllers are trained separately, the agent performance we get exceeds that of the base controller. With the combination of controllers, our agent's performance improves even more. The success rate is higher than the two base controllers and the model trained by them respectively, and the error rate is lower than both of them, with the average completion step size between base controller 0 and base controller 1. It can be seen that when using an ensemble of base controllers, our algorithm can learn the advantages of each base controller at the same time. The performance is better than using a single controller, which reflects the superiority of our algorithm. However, residual RL\cite{2019residual} can only use one base controller.

\begin{figure*}[h]
  \centering
  \subfigure[Stacking]{\includegraphics[width=0.31\textwidth]{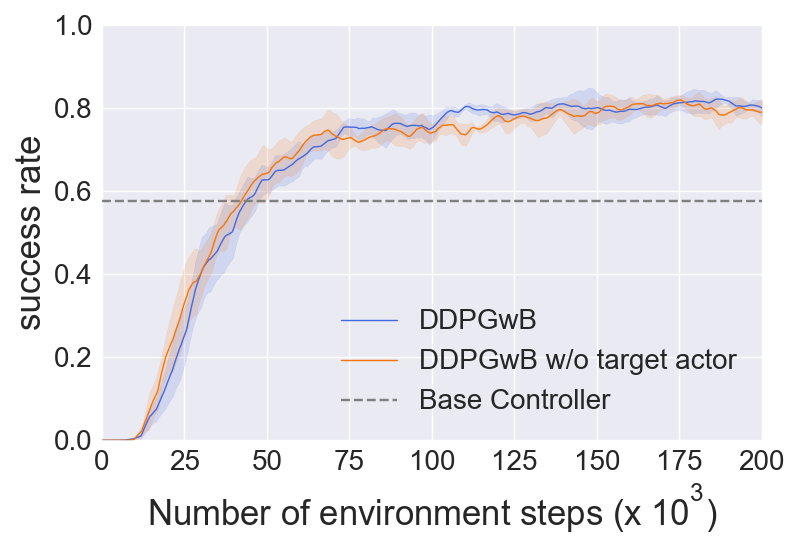}}
  \subfigure[Block-cup]{\includegraphics[width=0.31\textwidth]{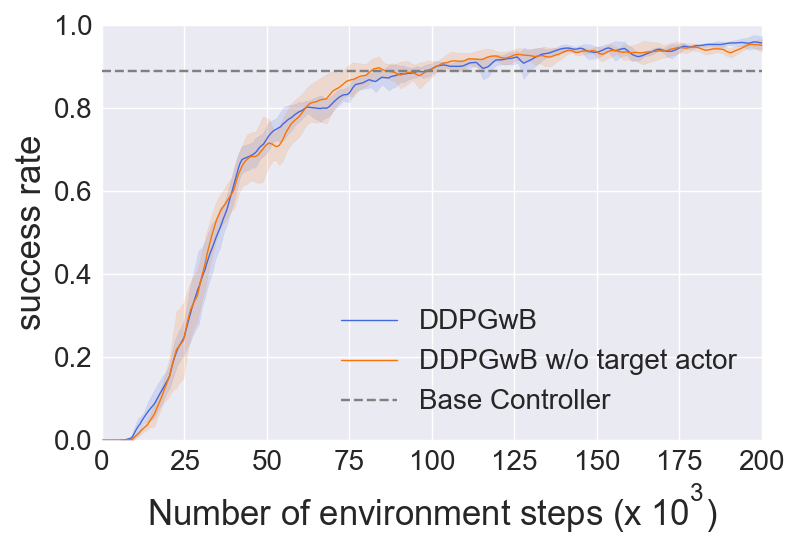}}
  \subfigure[Cup-cup]{\includegraphics[width=0.31\textwidth]{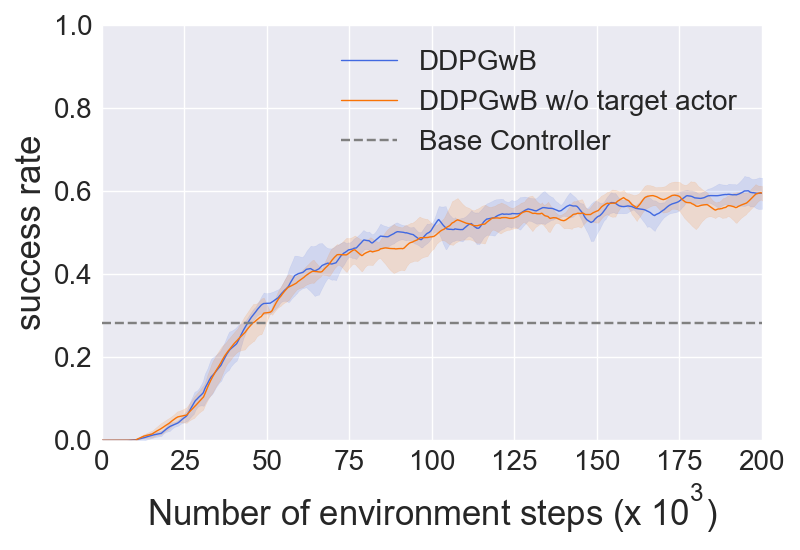}}
  \caption{Effect of abandoning the use of target actor networks. Without the target actor network, the performance of our algorithm is not significantly reduced.}
  \label{fig:notargactor}
\end{figure*}

\paragraph{Abandon the Target Actor}We further validate our algorithm by abandoning the use of a target actor network, which is an unusual ablation since DDPG needs both target actor and target critic networks for stabilized learning. As shown in Fig. \ref{fig:notargactor}, abandoning the use of a target actor network does not create drastic decay to our algorithm's performance, and policies are still stably learned outperforming base controllers, which highlights that our algorithm greatly increases the stability of actor. 

\section{CONCLUSION}
In this paper, we propose DDPGwB, an algorithm that utilizes base controllers to efficiently and safely learn challenging sparse-reward robot tasks. DDPGwB incorporates base controllers into stages of exploration, Q-value estimation as well as policy update, and learns state-based or vision-based policies that exceed the performances of base controllers. At the same time, our algorithm can be extended to an ensemble of base controllers to concentrate the advantages of each controller. The method proposed in our research may facilitate DRL applications in industrial robot systems where traditional non-DRL powered controllers are pervasive. 

One interesting future direction is to incorporate more base controllers for one single task, or to use a parameterized base controller distribution. Such a distribution might facilitate diverse control strategies and in turn aid exploration of the agent. Another possible direction is to use Task and Motion Planning algorithms \cite{dantam2016incremental} to automatically synthesize base controllers for tasks requiring multi-step planning.

\ifCLASSOPTIONcaptionsoff
  \newpage
\fi

\bibliographystyle{IEEEtran}  
\bibliography{IEEEabrv,bare_jrnl} 

\begin{IEEEbiography}[{\includegraphics[width=1in,height=1.25in,clip,keepaspectratio]{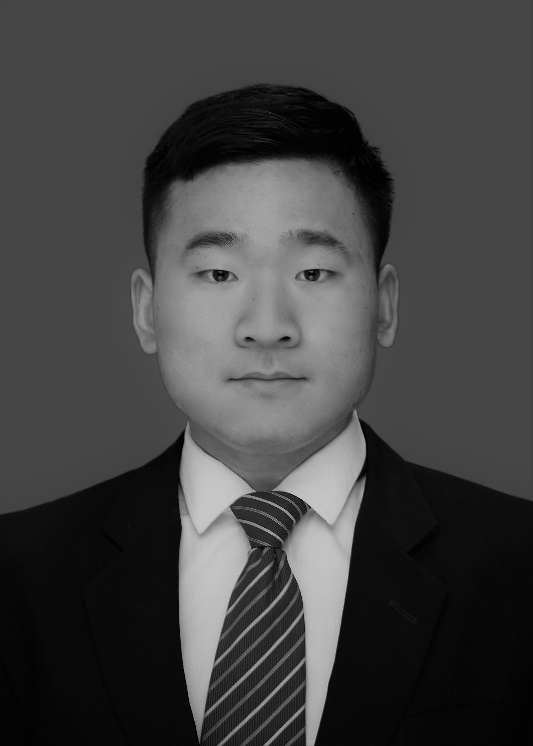}}]{Guangming Wang}
received the B.S. degree from Department of Automation from Central South University, Changsha, China, in 2018. He is currently pursuing the Ph.D. degree in Control Science and Engineering with Shanghai Jiao Tong University. His current research interests include robot learning and computer vision, in particular, deep reinforcement learning for robot control.
\end{IEEEbiography}

\begin{IEEEbiography}[{\includegraphics[width=1in,height=1.25in,clip,keepaspectratio]{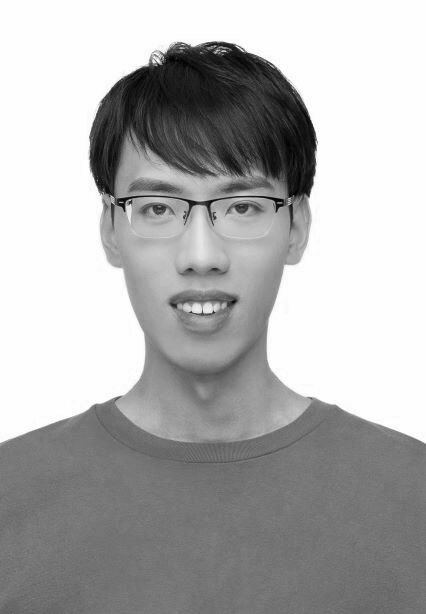}}]{Minjian Xin}
 received his Bachelor's Degree from the Department of Automation, Shanghai Jiao Tong University. He is currently pursuing the Master's Degree in Computer Science at University of California San Diego. His latest research interests include deep reinforcement learning and robot control.
\end{IEEEbiography}

\begin{IEEEbiography}[{\includegraphics[width=0.9in,height=1.3in,clip]{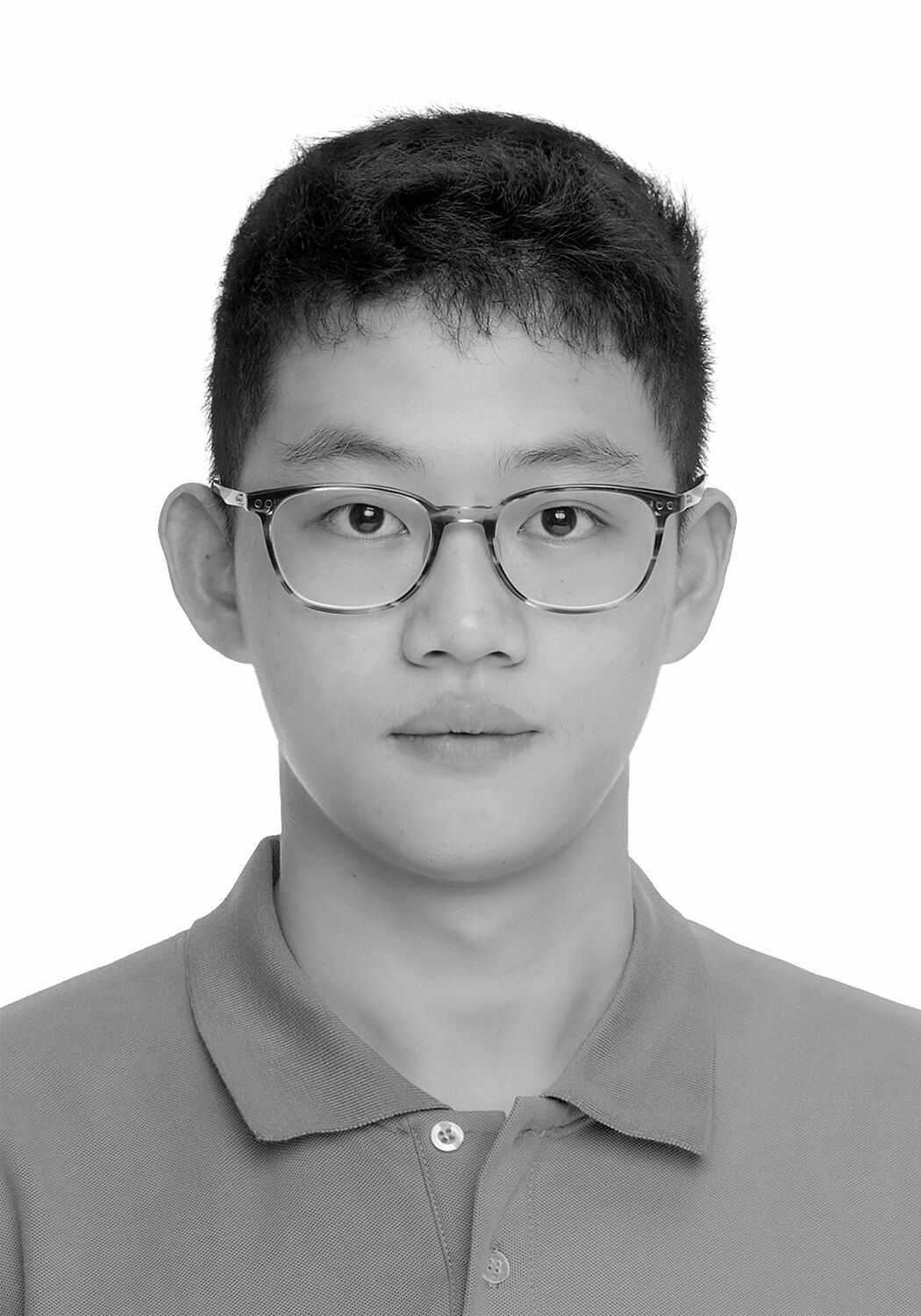}}]{Wenhua Wu}
is currently pursuing the B.S. degree in Department of Automation, Shanghai Jiao Tong University. His latest research interests include deep reinforcement learning and robot control. 
\end{IEEEbiography}
\begin{IEEEbiography}[{\includegraphics[width=1in,height=1.25in,clip,keepaspectratio]{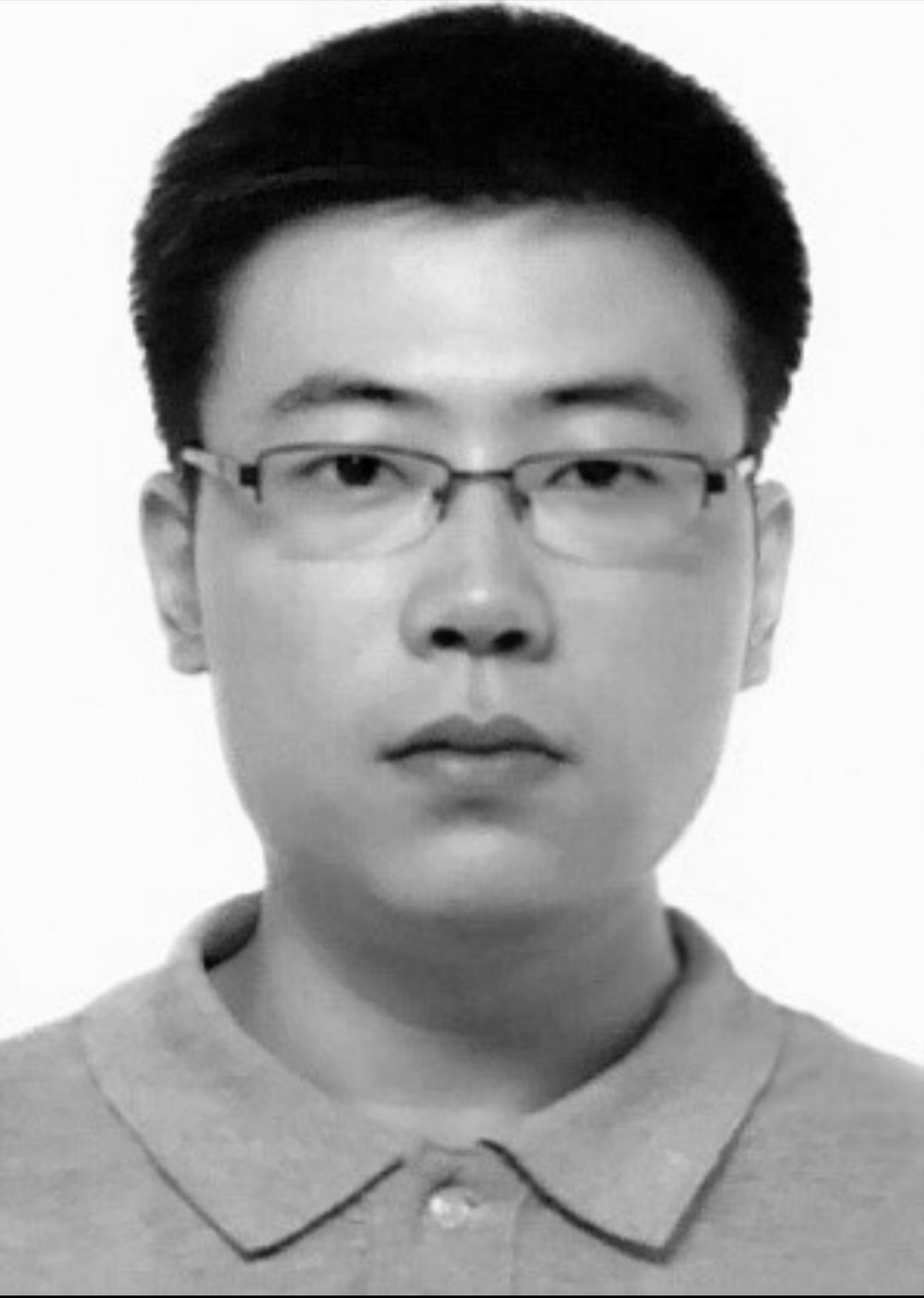}}]{Zhe Liu} received his B.S. degree in Automation from Tianjin University, Tianjin, China, in 2010, and Ph.D. degree in Control Technology and Control Engineering from Shanghai Jiao Tong University, Shanghai, China, in 2016. From 2017 to 2020, he was a Post-Doctoral Fellow with the Department of Mechanical and Automation Engineering, The Chinese University of Hong Kong, Hong Kong. He is currently a Research Associate with the Department of Computer Science and Technology, University of Cambridge. His research interests include autonomous mobile robot, multirobot cooperation and autonomous driving system. 
\end{IEEEbiography}

\begin{IEEEbiography}[{\includegraphics[width=1in,height=1.25in,clip,keepaspectratio]{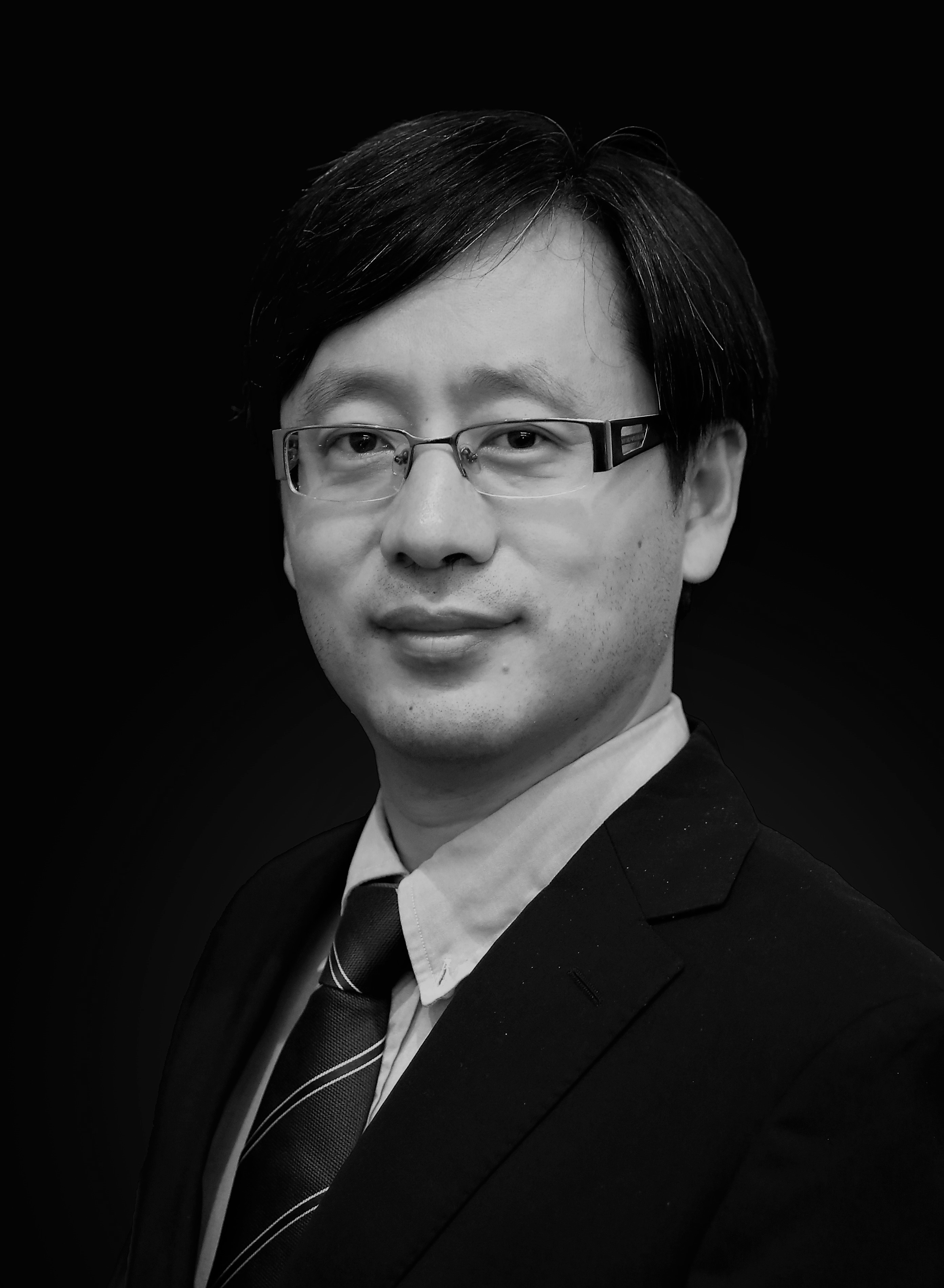}}]{Hesheng Wang}
received the B.Eng. degree in electrical engineering from the Harbin Institute of Technology, Harbin, China, in 2002, and the M.Phil. and Ph.D. degrees in automation and computer-aided engineering from The Chinese University of Hong Kong, Hong Kong, in 2004 and 2007, respectively. He is currently a Professor with the Department of Automation, Shanghai Jiao Tong University, Shanghai, China. His current research interests include visual servoing, service robot, computer vision, and autonomous driving. 
Dr. Wang is an Associate Editor of IEEE Transactions on Automation Science and Engineering, IEEE Robotics and Automation Letters, Assembly Automation and the International Journal of Humanoid Robotics, a Technical Editor of the IEEE/ASME Transactions on Mechatronics, an Editor of Conference Editorial Board (CEB) of IEEE Robotics and Automation Society. He served as an Associate Editor of the IEEE Transactions on Robotics from 2015 to 2019. He was the General Chair of the IEEE RCAR 2016, and the Program Chair of the IEEE ROBIO 2014 and IEEE/ASME AIM 2019. He is the General Chair of IEEE ROBIO 2022.

\end{IEEEbiography}

\end{document}